\documentclass{fairmeta}
\usepackage[most]{tcolorbox}

\usepackage{amsmath}
\usepackage{amssymb}
\usepackage{adjustbox}
\usepackage{wrapfig}

\usepackage{cuted}
\usepackage{capt-of}

\title{ADAPT: Analytical Disturbance-Aware Policy Training for Humanoid Locomotion}

\author[1, *]{Bofan Lyu}
\author[1, *]{\href{https://jiajindou.github.io/}{\textcolor{black}{Jindou Jia}}}
\author[1]{Kuangji Zuo}
\author[1]{Yanshuo Lu}
\author[1]{Shijia Han}
\author[1]{\href{https://reagan1311.github.io/}{\textcolor{black}{Gen Li}}}
\author[1]{\href{https://ma-boyu.github.io/}{\textcolor{black}{Boyu Ma}}}
\author[1]{\href{https://lorenzo-0-0.github.io/}{\textcolor{black}{Jingliang Li}}}
\author[1]{\href{https://genglibot.github.io/Sam/}{\textcolor{black}{Geng Li}}}
\author[1,\dagger]{\href{https://marsyang.site/}{\textcolor{black}{Jianfei Yang}}}

\affiliation[1]{MARS Lab, Nanyang Technological University}

\contribution[*]{Equal contribution}
\contribution[\dagger]{Corresponding authors}

\abstract{
Humanoids deployed in human-centered environments must handle force-interactive tasks, where external contacts introduce unexpected disturbances that disrupt locomotion accuracy and stability. Existing learning-based approaches rely on broad domain randomization, task-specific force objectives, or learning-based force estimators from motion history, each of which compromises accuracy, task transferability, or out-of-distribution (OOD) robustness. We present \textbf{A}nalytical \textbf{D}isturbance-\textbf{A}ware \textbf{P}olicy \textbf{T}raining (\textbf{ADAPT}), a framework that equips humanoid policies with a physically grounded disturbance observer. The core of ADAPT is an analytical whole-body disturbance observer that estimates residual force/torque online with the accessible robot dynamics, without requiring force/torque sensors. Fed directly into the policy, the estimated disturbances give the humanoid an explicit, physics-derived sense of external force/torque that can generalize across diverse unseen scenes. Experiments on a Unitree G1 humanoid show that ADAPT achieves accurate disturbance prediction and stronger robustness than a proprioception-only baseline under torso perturbations, standing pushes, and asymmetric hand payloads, with improved velocity tracking even on OOD disturbances. Moreover, ADAPT enables penalizing inferred disturbances at lower-body joints to encourage lighter locomotion.
}

\correspondence{Jianfei Yang: \email{jianfei.yang@ntu.edu.sg}, Bofan Lyu: \email{lyub0002@e.ntu.edu.sg}.}
\metadata[Project]{\url{https://blyu413.github.io/adapt-locomotion/}.}

\begin{document}

\maketitle

\section{Introduction}
Recent progress in learning-based control has greatly advanced humanoid locomotion in the real world~\citep{radosavovic2024RealworldHumanoidLocomotion,liao2025BeyondMimicMotionTracking,luo2025SONICSupersizingMotion, yuan2026roboforgephysicallyoptimizedtextguided}. As humanoids move toward deployment in human-centered environments, they inevitably encounter force-interactive tasks, in which robots must exert forces while adapting to the resulting reaction forces from the environment. Examples include transporting payloads~\citep{zhang2025FALCONLearningForceAdaptive,du2025LearningHumanHumanoidCoordination}, pushing loaded carts~\citep{li2026HAICHumanoidAgile}, opening doors~\citep{xue2025OpeningSimtoRealDoor}, and interacting compliantly with humans~\citep{chen2025CHIPAdaptiveCompliance,lu2025GentleHumanoidLearningUpperbody}. These physically interactive tasks inevitably introduce a wide range of unseen uncertainties that disrupt the accuracy and stability of the robot's locomotion. How to clearly characterize these uncertainties and effectively overcome them becomes increasingly critical for reliable humanoid deployment.


Existing learning-based methods have explored several ways to handle contact-rich interactive tasks. One line of work aims to improve sim-to-real robustness by exposing the policy to randomized external forces and internal dynamics variations during training~\citep{radosavovic2024RealworldHumanoidLocomotion,long2024LearningHInfinityLocomotion,zhang2025TrackAnyMotions}. Another line of work explicitly incorporates force perception into loco-manipulation training for task-specific objectives such as exerting desired forces~\citep{portela2024LearningForceControl}, producing compliant responses~\citep{chen2025CHIPAdaptiveCompliance, lu2025GentleHumanoidLearningUpperbody}, or enabling force-adaptive control~\citep{zhang2025FALCONLearningForceAdaptive,xu2025FACETForceAdaptiveControl}. A third direction uses learning-based force estimators that infer interaction forces from proprioceptive histories, and then feed the predicted force-related signal to the controller to form force-conditioned policies without additional force/torque sensors~\citep{zhi2025learningunifiedpolicy}. However, these methods either rely on broad randomization that sacrifices tracking accuracy, require task-specific design, or generalize poorly to out-of-distribution (OOD) disturbances.


\begin{figure*}
    \centering
    \vspace{-13pt}
    \includegraphics[width=0.9\linewidth]{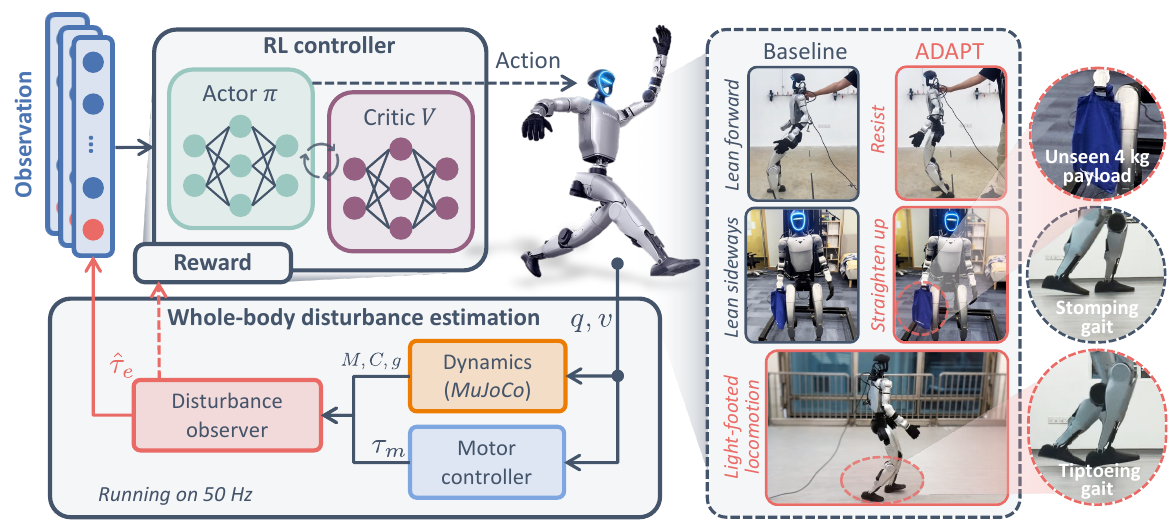}
    \vspace{-5pt}
    \caption{\textbf{Overview of ADAPT.} ADAPT augments RL-based humanoid control with an analytical disturbance observer that infers whole-body disturbance torques from accessible robot dynamics, proprioception, and expected motor torques. The estimated disturbances are fed into the policy as an explicit observation and used for reward shaping. This enables emergent force-aware behaviors, including push resistance, payload compensation, and light-footed locomotion.}
    \label{fig:method_overview}
    \vspace{-10pt}
\end{figure*}

This motivates us to seek a more general and minimally policy-invasive disturbance-aware strategy for improving the robustness and generalization of humanoid policies. Physics-informed learning has been increasingly shown to be an effective path toward more robust and generalizable policies~\citep{lutter2019deep, jia2026Physics}. Following this line of thought, we ask whether prior physical knowledge of the humanoid itself can serve as a source of disturbance awareness, one that informs policy learning and, being grounded in physics rather than data, generalizes beyond the training distribution. Analytical disturbance estimation has been extensively explored in classical robot control, exemplified by nonlinear disturbance observers~\citep{chen2000NonlinearDisturbanceObserver,jia2025FORESEERR} and $L_1$ adaptive estimation~\citep{han2009FromPIDActive}. Yet, integrating such analytical estimates into Reinforcement Learning (RL)-based humanoid policies to improve robustness and generalization remains an open question~\citep{lim2025MOBNetLimbmodularizedUncertainty}.

In this work, we introduce the \textbf{A}nalytical \textbf{D}isturbance-\textbf{A}ware \textbf{P}olicy \textbf{T}raining (\textbf{ADAPT}) framework for humanoid locomotion, which augments prevailing RL-based policy with explicit whole-body disturbance prediction during training. Specifically, instead of requiring the policy to infer external forces/torques only from motion histories~\citep{portela2024LearningForceControl,zhi2025learningunifiedpolicy}, we estimate unexpected forces/torques online with accessible physical knowledge and feed them directly into the policy. This gives the policy a structured and physically grounded force-related perception channel for disturbance-aware humanoid control. Notably, the analytical disturbance estimation also enables task-specific reward shaping, such as penalizing foot-joint disturbances to encourage lighter locomotion. On the technical side, we further integrate a Fast-LIO~\citep{xu2022fast} based odometry module for accurate root linear velocity, which, to our knowledge, is among the earliest uses of LIO-based root velocity estimation as policy inputs in learning-based humanoid whole-body control.

We validate ADAPT in both simulation and real-world experiments on a Unitree G1 humanoid. The results show that the proposed whole-body disturbance observer provides a consistent and reliable disturbance signal. With this prediction as policy input, ADAPT improves robustness against external forces, achieving better velocity tracking and smaller directional drift than a proprioception-only baseline. More importantly, the policy also shows an emergent force-aware response. Under asymmetric hand payloads never encountered during training, it actively adjusts the whole-body posture to compensate for the load. Beyond direct policy feedback, the same disturbance prediction can also support contact-aware reward shaping and behavior analysis.

\section{Related works}
\textbf{Force-aware policy learning.}
Handling interaction forces is central to contact-rich humanoid and legged control, where robots must apply, resist, or yield to external contacts while remaining stable. Existing learning-based methods address this challenge in several ways. Some methods improve robustness by training policies with randomization of internal dynamics~\citep{radosavovic2024RealworldHumanoidLocomotion, zhang2025TrackAnyMotions} and external disturbances~\citep{zhang2025TrackAnyMotions, long2024LearningHInfinityLocomotion}. Other methods introduce force interaction through task-specific objectives and rewards, enabling tasks such as payload transport~\citep{zhang2025FALCONLearningForceAdaptive}, door opening~\citep{xue2025OpeningSimtoRealDoor}, high-intensity pulling~\citep{li2025ThorHumanLevelWholeBody}, compliant interaction~\citep{chen2025CHIPAdaptiveCompliance,lu2025GentleHumanoidLearningUpperbody,hou2025adaptivecompliancepolicy}, and collaborative carrying~\citep{du2025LearningHumanHumanoidCoordination}. Another line of work estimates or predicts force-related quantities from proprioceptive histories for direct force control~\citep{portela2024LearningForceControl}, unified position and force control~\citep{zhi2025learningunifiedpolicy}, or disturbance compensation~\citep{zhang2025DisturbanceAwareAdaptiveCompensation}. These approaches handle external forces by broadening the training distribution, embedding interaction requirements into task objectives, or learning force-related representations from proprioceptive histories. In contrast, ADAPT leverages a physics-derived disturbance prediction with built-in generalization.

\textbf{Sensorless disturbance estimation.}
Some methods infer object physical properties or task feasibility from internal robot measurements and a priori physical knowledge, without relying on joint force/torque sensors~\citep{lao2023learningbasedapproach,han2020CanLiftIt}. They have been used for disturbance rejection, sensorless torque reasoning, fault detection, and collision monitoring for various robots~\citep{chen2000NonlinearDisturbanceObserver,chen2004DisturbanceObserverBased,deluca2003ActuatorFailureDetection,haddadin2017RobotCollisionsSurvey}. For humanoids, external torque estimation is more challenging due to high-dimensional, strongly nonlinear dynamics. MOB-Net~\citep{lim2025MOBNetLimbmodularizedUncertainty} uses a momentum observer to estimate external contact wrenches, primarily for wrench feedback, collision detection, and collision reaction. BeyondMimic~\citep{liao2025BeyondMimicMotionTracking} combines the momentum observer with a Kalman filter for onboard full-state estimation during humanoid deployment. These estimated forces are commonly used for monitoring, identification, or model-based feedback. However, directly incorporating such analytical disturbance estimates into the learned policy as an explicit observation modality remains largely unexplored.

\section{Methodology}
\subsection{Overview of ADAPT}

\textbf{Problem formulation.} The whole-body dynamics of a humanoid~\citep{nenchev2018humanoid} can be formalized as
\begin{align}
    M (q)\dot v + C (q,v)v + g (q) = \tau_m + \tau_e,
    \label{eq:humanoid_dynamics} 
\end{align}
where $q \in \mathbb{R}^{n_j+7}$ and $v \in \mathbb{R}^{n_j+6}$ are the generalized coordinate and velocity, and $n_j$ denote the number of actuated joints in the humanoid. The dimension mismatch comes from the floating-base representation: $q$ includes a 3D base position and a 4D unit quaternion, whereas $v$ includes 3D base linear and 3D angular velocities. $M (q)$ is the nominal inertia matrix, $C (q,v)v$ is the nominal \textit{Coriolis} and centrifugal term, $g (q)$ is the nominal gravity term, $\tau_m$ is the generalized actuation force induced by joint motors, and $\tau_e$ denotes the generalized disturbance. In practice, $\tau_e$ captures the combined effect of external contacts (e.g., payloads, environmental reactions, human interaction) and internal model mismatch (e.g., friction, inertial uncertainty) on the robot's generalized motion.

Direct access to $\tau_e$, however, is difficult in practice. Humanoid robots are rarely equipped with a whole-body force/torque sensor. Meanwhile, directly computing $\tau_e$ through Eq.~\eqref{eq:humanoid_dynamics} requires the generalized acceleration $\dot v$, whose estimation is highly susceptible to sensor noise and numerical differentiation errors. Existing force-aware policies often obtain disturbance cues through task-specific curricula~\citep{chen2025CHIPAdaptiveCompliance} or learned force estimators from motion history~\citep{zhi2025learningunifiedpolicy}, which makes the resulting force representation dependent on the disturbance distribution specified during training and inevitably degrades under out-of-distribution forces.

\textbf{Our solution.} Our ADAPT framework addresses this problem by closing the loop between analytical disturbance estimation and policy control. As illustrated in Fig.~\ref{fig:method_overview}, ADAPT consists of two main modules: an analytical disturbance estimator and an RL-based whole-body controller. The disturbance estimator, detailed in Sec.~\ref{sec:whole_body_observer}, evaluates the nominal robot dynamics from the current robot state and actuator command, and estimates the whole-body disturbance. The learning-based controller, detailed in Sec.~\ref{sec:policy_network}, takes this estimation as a force-aware proprioceptive signal and feeds it back to the policy as part of its observation or even the upper-level reward. With this feedback channel, the learned controller can condition its whole-body response on the inferred disturbance, enabling disturbance-aware humanoid adaptation.

\subsection{Whole-body disturbance observer}
\label{sec:whole_body_observer}
Derived from Eq.~\eqref{eq:humanoid_dynamics}, we can easily get a representation of whole-body disturbance through $\tau_e = M (q)\dot v + C (q,v)v + g (q) - \tau_m$. However, directly calculating $\dot v$ is impractical, as it is usually unavailable from the sensory system and noisy under numerical differentiation. The disturbance observer avoids this noisy acceleration term by using the generalized momentum, i.e., $p = M (q)v$.
Differentiating the generalized momentum and using the standard \textit{Christoffel} convention, $\dot M (q)=C (q,v)+C ^{\top}(q,v)$~\citep{deluca2003ActuatorFailureDetection}, gives
\begin{align}
    \dot p
    &=
    \dot M  v + M  \dot v
    =
    C  v + C ^{\top}v + M \dot v.
    \label{eq:momo_pdot}
\end{align}
Substitute Eq.~\eqref{eq:humanoid_dynamics} into Eq.~\eqref{eq:momo_pdot} gives $\tau_e = \dot p - \beta (q,v) - \tau_m$, 
where $\beta (q,v) = C ^{\top}(q,v)v - g (q)$ for simplicity. Now, we can estimate $\tau_e$ using a first-order observer with positive-definite diagonal observer gain $K_o$, i.e., $\dot{\hat{\tau}}_e =K_o({{\tau}}_e - \hat{{\tau}}_e)$. To avoid the use of the inaccessible $\dot p$, we introduce the auxiliary state $z=\hat{\tau}_e-K_op$ and obtain
\begin{align}
    \dot z
    &=
    -K_o\left(\beta +\tau_m+\hat{\tau}_e\right),
    \label{eq:momo_z_dynamics} \\
    \hat{\tau}_e
    &=
    z+K_op .
    \label{eq:momo_tau_hat}
\end{align}
Note that the observer established in Eqs.\eqref{eq:momo_z_dynamics}-\eqref{eq:momo_tau_hat} only requires accessible signals including current proprioceptive state, control command, and nominal dynamics terms. 

As for the nominal dynamics, instead of hand-deriving, we obtain $M (q)$ and $\beta$ directly from a synchronized MuJoCo model, which is fast enough for large-scale parallel training and deployment. And $\tau_m$ is derived from an ideal motor controller. For efficiency, we approximate $\beta (q,v) = C ^{\top}(q,v)v - g (q)$ using the negative MuJoCo bias force $-q_{\mathrm{bias}}$, where $q_{\mathrm{bias}} = C (q,v)v + g (q)$. This approximation introduces a velocity-dependent error $\dot M (q)v$ in $\beta$, whose effect is shown to be negligible in practice (see Appx.~\ref{observer_approximation}).

The observer gain $K_o$ governs a fundamental trade-off. A larger $K_o$ accelerates convergence and improves estimation accuracy, but also amplifies high-frequency noise from measurement, especially critical on real hardware. To balance responsiveness and noise robustness, we apply a low-pass filter to the raw estimate $\hat{\tau}_e$ before passing it to downstream modules. Moreover, the linear floating-base residual contains a constant gravity bias in the inertial vertical direction. We remove this constant gravity bias and then rotate the floating-base residual into the root frame before feeding it to the policy. We denote the final resulting filtered residual as
\begin{equation}
    \bar{\tau}_e = 
    \left[\bar{\tau}^{\mathrm{lin}}_{e}, 
    \bar{\tau}^{\mathrm{ang}}_{e}, 
    \bar{\tau}^{\mathrm{jnt}}_{e}\right],
    \label{eq:momo_tau_final}
\end{equation}
where $\bar{\tau}^{\mathrm{lin}}_{e} \in \mathbb{R}^{3}$ and $\bar{\tau}^{\mathrm{ang}}_{e} \in \mathbb{R}^{3}$ are the linear and angular residuals of the floating base, expressed in the root frame of the humanoid. The term $\bar{\tau}^{\mathrm{jnt}}_{e} \in \mathbb{R}^{29}$ denotes the residual in the joint coordinates.

\subsection{Disturbance-aware policy}
\label{sec:policy_network}
With the analytical disturbance estimation, we next endeavor to design the RL policy. We use a multi-layer perceptron (MLP) as the whole-body controller. The policy is trained with Proximal Policy Optimization (PPO)~\citep{DBLP:journals/corr/SchulmanWDRK17} and outputs target joint positions, which are tracked by a low-level motor PD controller. The key modification in ADAPT is to augment the actor input with the analytical disturbance estimated by the dynamics-driven observer in Sec.~\ref{sec:whole_body_observer}. This analytical disturbance provides a structured force-related observation channel, enabling the policy to condition its response on explicit whole-body disturbance. See Fig.~\ref{fig:method_overview} for the policy structure.

\textbf{Actor-critic architecture.} We use an asymmetric \textit{actor-critic} architecture trained with PPO~\citep{DBLP:journals/corr/SchulmanWDRK17}. Both the actor and critic are three-layer MLPs with hidden dimensions $(512,256,128)$ and ELU activations. At timestep $k$, the actor observation $o^\pi_k$ consists of a five-step history from $k-4$ to $k$ of base linear velocity $v^b_k\in\mathbb{R}^3$, base angular velocity $\omega^b_k\in\mathbb{R}^3$, projected gravity $g^b_k\in\mathbb{R}^3$, joint position and velocity $q^j_k,\dot q^j_k\in\mathbb{R}^{n_j}$, velocity command $c_k\in\mathbb{R}^3$, and effort-scaled disturbance estimation $\rho_k\in\mathbb{R}^{6+n_j}$, together with the previous action $a_{k-1}\in\mathbb{R}^{n_j}$, resulting in a 554-dimensional actor observation.  Moreover, the critic receives the actor observation and additional privileged foot-contact information, including foot height $h_f$, foot air time $t_f^{\mathrm{air}}$, contact state $s_f$, and contact force $F_f$. These privileged observations are used only during training for value estimation.

\textbf{Training pipeline.} To allow the policy to gradually adapt to external disturbance and incorporate the disturbance observation, we train it with a two-stage curriculum. Stage 1 trains the policy to acquire basic locomotion in a disturbance-free setting, while still consuming the disturbance observation as input. Stage 2 introduces a disturbance curriculum with external forces applied to selected bodies, requiring the same policy to track velocity commands under increasing disturbance levels.

\textbf{Disturbance normalization.} For the disturbance observation, standard normalization is not well-suited to our two-stage force curriculum, as the disturbance magnitudes encountered in the two stages differ significantly in scale. We therefore use unified scales for the disturbance observation channels in two stages, i.e., $\rho = \left[{\bar{\tau}^{\mathrm{lin}}_{e}}/({m  g}), {\bar{\tau}^{\mathrm{ang}}_{e}}/({m  g l_{\mathrm{ref}}}), {\bar{\tau}^{\mathrm{jnt}}_{e}}/{\tau^{\max}_j}\right]$, where $m $ is the nominal robot mass, $g$ is the gravity magnitude, $l_{\mathrm{ref}}$ is a user-defined normalization constant, and $\tau^{\max}$ is the actuator effort limit of each joint. This representation incorporates joint-dependent actuation capacity by expressing each joint's residual relative to its corresponding effort limit. We apply this unified normalization to the disturbance channels and standard observation normalization to the remaining policy inputs. Appx.~\ref{effort_scale_ablation} shows the effectiveness of this scaling strategy.

\textbf{Training reward.} Within the ADAPT framework, in addition to the standard rewards for command tracking~\citep{zakka2026mjlablightweightframeworkgpuaccelerated}, ADAPT supports task-specific reward terms defined on the estimated disturbance. Here, we penalize the leg-joint disturbance estimates to encourage lighter, lower-impact locomotion. Let $\rho^s_i$ denote the scaled leg disturbance residuals of the $i$-th joint of leg $s\in\{L,R\}$ at a single timestep. For each leg, we define the residual envelope as $e^s=\sqrt{\frac{1}{6}\sum_{i=1}^{6}\left(\rho^s_i\right)^2}$.

During training, this value is computed at each timestep, and we use $e_k=\max(e^L_k,e^R_k)$ as the lower-body residual envelope at timestep $k$. Let $\mathcal{H}_k=\{e_{k-H+1},\ldots,e_k\}$ denote a rolling window of length $H$. We compute the high-tail average $\eta_k=\mathrm{mean}\left(\mathrm{Top}_{10\%}(\mathcal{H}_k)\right)$ as the mean of the largest 10\% values in $\mathcal{H}_k$, and the peak value $e^{\mathrm{peak}}_k=\max_{e\in\mathcal{H}_k}$. We define two penalty as $\phi_{\eta,k} = \max\!\left(0,\frac{\eta_k-\theta_{\eta}}{\delta_{\eta}}\right)$ and $\phi_{\mathrm{peak},k} = \max\!\left(0,\frac{e^{\mathrm{peak}}_k-\theta_{\mathrm{peak}}}{\delta_{\mathrm{peak}}}\right)$. The light-step reward is

\begin{equation}
\begin{aligned}
r^{\mathrm{light}}_k
&= -\chi_k
\left(
\lambda_{\eta}\phi_{\eta,k}
+\lambda_{\mathrm{peak}}\phi_{\mathrm{peak},k}
\right).
\end{aligned}
\label{eq:light_step_reward}
\end{equation}


where $\chi_k\in\{0,1\}$ is a command mask that enables the penalty for locomotion commands and disables it for zero commands, $\lambda_{\eta}$ and $\lambda_{\mathrm{peak}}$ are penalty weights, $\theta_{\eta}$ and $\theta_{\mathrm{peak}}$ are activation thresholds, and $\delta_{\eta}$ and $\delta_{\mathrm{peak}}$ are margin parameters that scale the excess above the thresholds. These terms penalize high-impact lower-body residual peaks, while preserving the support forces needed for walking. Other instantiations, such as penalizing hand-joint disturbances to encourage compliant pushing or pulling, are left for future exploration.

\subsection{Practical consideration}
\textbf{FAST-LIO usage.}
The root linear velocity is an important observation for the disturbance observer, but it is not directly available in the current humanoid community. Directly integrating IMU acceleration is unreliable due to drift. We here employ FAST-LIO~\citep{xu2022fast} to obtain the robot motion during real-world deployment. Since the LiDAR is mounted on the torso, we convert the raw observation to the root frame before using it in the disturbance observer and policy. Let $v^{L}_{L}\in\mathbb{R}^3$ be the raw LiDAR linear velocity from FAST-LIO expressed in the LiDAR local frame. Given the current joint position $q_{\mathrm{jnt}}$ and the rotation matrix $R_{RL}(q_{\mathrm{jnt}})\in SO(3)$ that maps vectors from the LiDAR frame to the root frame, the LiDAR velocity expressed in the root frame is $v^{R}_{L} = R_{RL}(q_{\mathrm{jnt}}) v^{L}_{L}$. We then compute the root linear velocity as $v^{R}_{\mathrm{root}} = v^{R}_{L} - J_{L}(q_{\mathrm{jnt}})\dot{q}_{\mathrm{jnt}}$, 
where $v^{R}_{\mathrm{root}}$ is the root linear velocity expressed in the root frame, $J_{L}(q_{\mathrm{jnt}})\in\mathbb{R}^{3\times n_j}$ is the LiDAR linear Jacobian with respect to the actuated joints, expressed in the root frame, and $\dot{q}_{\mathrm{jnt}}$ is the measured joint velocity. The term $J_{L}(q_{\mathrm{jnt}})\dot{q}_{\mathrm{jnt}}$ represents the relative linear velocity between the LiDAR and the root.


\section{Experimental Results}
We evaluate ADAPT in both simulation and real-world experiments on a Unitree G1 humanoid. The experiments are designed to answer three questions: (1) Does the disturbance observer provide accurate disturbance estimation during humanoid locomotion? (2) Does conditioning the policy on the analytical disturbance improve robustness and generalization to external perturbations? (3) Does using the analytical disturbance for reward reshaping bring additional benefits? The experimental setup and parameter setting are detailed in Appx.\ref{experiment_setup}.

\subsection{Estimation performance of disturbance observer}
\begin{figure*}[t]
    \centering
    \vspace{-10pt}
    \includegraphics[width=0.9\linewidth]{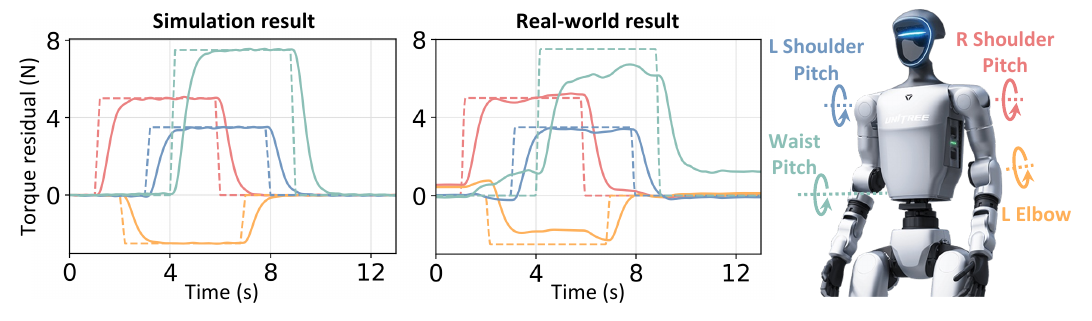}
    \vspace{-14pt}
    \caption{\textbf{Performance of the disturbance observer on the Unitree G1}. The robot walks forward while known disturbance torques (dashed lines) are injected into the left shoulder pitch, left elbow, right shoulder pitch, and waist pitch joints. Solid lines show the observer estimation in simulation (\textit{left}) and real-world (\textit{middle}).}
    \label{fig:performance_observer}
    \vspace{-10pt}
\end{figure*}

The observer for estimating whole-body disturbance is the core component of our ADAPT framework, as it provides the policy with an explicit force-aware signal for physical interaction. Its performance directly affects the reliability of the learning-based controller built upon it. In both simulation and real test, the robot is commanded to walk forward with a fixed velocity command, $v_x = 0.8~\mathrm{m/s}$. The observer output is not provided to the policy observation and does not affect the control action. Therefore, the measured residuals reflect the standalone estimation behavior of the observer. During the rollout, known disturbance torques are directly injected into selected joints: left shoulder pitch, left elbow, right shoulder pitch, and waist pitch, as shown in the right of Fig.~\ref{fig:performance_observer}. The goal of this experiment is to validate whether the observer remains stable and reliable during locomotion.

The test results are shown in Fig.~\ref{fig:performance_observer}. In simulation, the observer closely tracks the injected disturbance profiles across multiple joints, and the estimated residuals remain well isolated among different joint channels. In the real-world test, the observer still captures the onset and relative magnitude of the applied disturbance in each channel, indicating consistent observer behavior between simulation and hardware. Compared with simulation, the observer shows a larger estimation error in the real world. This degradation reflects additional unknown real-world disturbances beyond the injected joint torque, such as joint friction and motor modeling error.

\subsection{Control performance of disturbance-aware policy}

\begin{figure*}[t]
    \centering
    \vspace{-10pt}
    \includegraphics[width=0.95\linewidth]{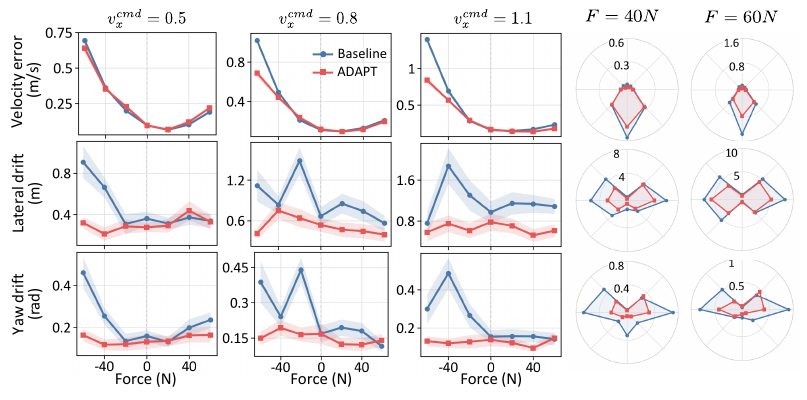}
    \vspace{-12pt}
    \caption{\textbf{Torso pulling test in simulation.}
    \textit{Left:} Velocity tracking error, lateral drift, and yaw drift under sagittal torso forces across three command velocities. \textit{Right:} Radar plots summarize results across disturbance directions.}
    \label{fig:experiment_policy_sim_torsopull}
    \vspace{-10pt}
\end{figure*}

The observer evaluation above shows that the disturbance observer provides a reliable estimation of joint-space disturbances. We next ask whether this estimated signal can be effectively used by a learned policy. To answer this question, we evaluate ADAPT under diverse velocity commands and external force perturbations. Perturbations vary in location and magnitude. We compare ADAPT with a baseline that uses the same policy architecture and training setup but does not receive the explicit disturbance feedback. The comparison is made in both simulation and real-world experiments. We adopt several evaluation metrics, including velocity tracking error $e_{v_x}$, lateral drift $d_y$, and yaw drift $d_{\psi}$. Detailed definitions are provided in Appx.~\ref{Evaluation_criterion}.

\textbf{Torso pulling test.} We first evaluate ADAPT in simulation by applying torso forces in the sagittal plane. The magnitude of forces ranges from $0$ N to $60$ N, where the maximum force magnitude used during training is $40$ N. Fig.~\ref{fig:experiment_policy_sim_torsopull} shows the results under sagittal torso forces across different walking speeds. Compared with the baseline, ADAPT achieves lower forward velocity tracking error, lateral drift, and heading drift, especially under stronger perturbations. This shows that the explicit disturbance feedback helps the policy compensate for external forces while maintaining the commanded locomotion behavior.

We further evaluate the velocity tracking task under torso perturbations applied from different horizontal directions. The radar maps in Fig.~\ref{fig:experiment_policy_sim_torsopull} present two force magnitudes, 40~N, which lies within the training range, and 60~N, which exceeds it. ADAPT outperforms the baseline, achieving better locomotion under both in-distribution and OOD disturbance magnitudes. These results indicate that the explicit disturbance estimation serves as an effective feedback for the policy, improving its robustness and generalization.

\begin{figure}
    \centering
    \includegraphics[width=0.95\linewidth]{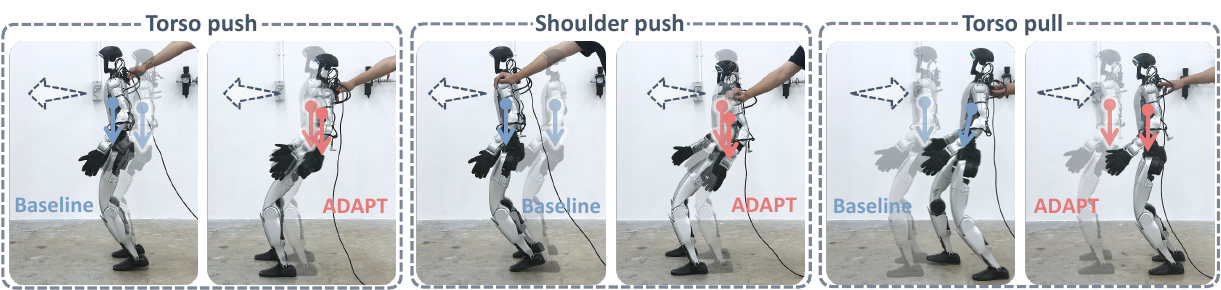}
    \vspace{-6pt}
    \caption{\textbf{Stance pulling test.} The policy trained under the ADAPT framework exhibits whole-body posture adjustment to resist external force applied on torso.}
    \vspace{-10pt}
    \label{fig:stance_pulling}
\end{figure}

\begin{wrapfigure}{r}{0.55\textwidth}
    \vspace{-15pt}
    \centering
    \includegraphics[width=\linewidth]{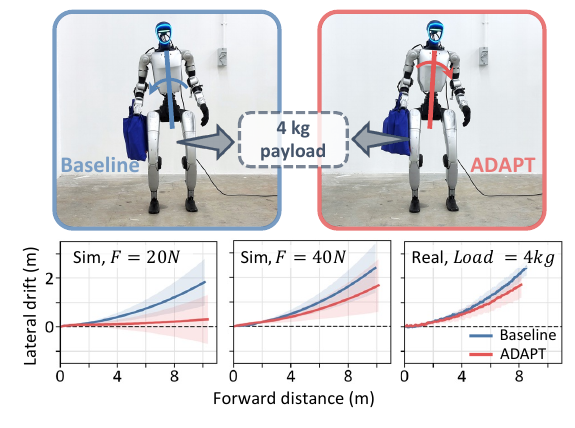}
    \vspace{-25pt}
    \caption{\textbf{Asymmetric hand loading test.} \textit{Upper}: postural response with an OOD 4 kg payload on the right hand. \textit{Lower}: lateral drift versus forward distance under one-sided hand loads, in both simulation and real tests.}
    \label{fig:pose_compare_ee_load}
    \vspace{-15pt}
\end{wrapfigure}

\textbf{Stance pulling test.}
On the real robot, we conduct a standing perturbation test to evaluate the policy response to external forces. The robot receives a zero-velocity command, while an operator applies external forces to its torso and shoulder. As presented by snapshots in Fig.~\ref{fig:stance_pulling}, compared with the baseline, the ADAPT-trained policy shows stronger resistance to the applied forces and adjusts its whole-body posture to counteract the perturbation. This indicates that the explicit disturbance estimation provides useful feedback for improving whole-body robustness during physical interaction.

\textbf{Asymmetric hand loading test.} We then evaluate ADAPT under asymmetric hand loading. In simulation, we apply a constant downward force of $10$--$40,\mathrm{N}$ to one hand. On the real robot, we reproduce this condition by attaching weights of 1 to 4 kg to the same hand, where the $4\,\mathrm{kg}$ payload lies outside the training distribution. The added payloads create downward loads comparable to the simulated setting. Under these asymmetric loads, ADAPT and the baseline exhibit markedly different postural responses. As shown in Fig.~\ref{fig:pose_compare_ee_load}, the proprioception-only baseline is pulled toward the loaded side, resulting in a visibly unbalanced upper-body posture. In contrast, ADAPT tilts the torso away from the loaded side to compensate for the moment caused by the load. We further examine walking robustness under asymmetric loading. The curves in Fig.~\ref{fig:pose_compare_ee_load} show that the baseline exhibits increasing lateral drift as the hand load becomes larger. In contrast, ADAPT maintains a straighter walking trajectory across all load magnitudes, indicating better compensation.

\subsection{Reward shaping for lighter locomotion}
Beyond using the disturbance estimation as a direct policy input, we also examine whether it can be used for reward shaping. We focus on foot touchdown as an illustrative test case, considering that abrupt touchdown often produces short disturbance peaks in the lower-body residuals. This allows us to test whether rewards derived from joint-space disturbance estimation can encourage lighter locomotion behavior. The reward shaping strategy is formalized in Eq.~\eqref{eq:light_step_reward}.

\begin{figure}[t]
    \centering
    \includegraphics[width=0.9\linewidth]{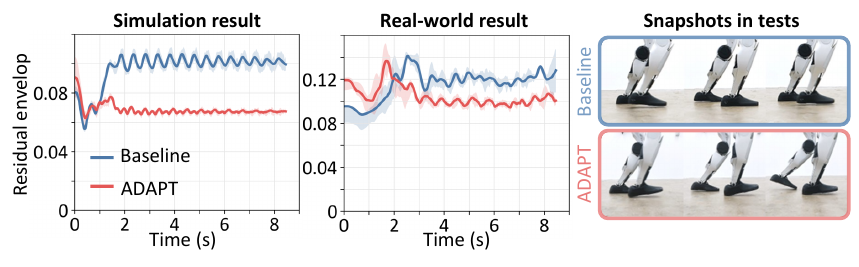}
    \vspace{-10pt}
    \caption{\textbf{Light-step Reward Shaping.} Disturbance residual envelope over time in simulation (\textit{left}) and on the real robot (\textit{middle}). \textit{Right:} Snapshots of the touchdown phase, showing that ADAPT produces a \textit{tiptoeing} gait while the baseline exhibits a \textit{stomping} gait.}
    \vspace{-12pt}
    \label{fig:obs_reward_shaping}
\end{figure}

We evaluate the light-step reward in both simulation and real-world deployment using the scaled leg disturbance residual envelope, where lower values indicate lighter leg loading and smaller foot-ground impact. As shown in Fig.~\ref{fig:obs_reward_shaping}, the light-step policy maintains a consistently lower envelope than the baseline and reduces the periodic loading peaks. This indicates lighter foot contacts during walking. This phenomenon can also be revealed in snapshots of the touchdown phase. Overall, the simulation and real-robot results show that ADAPT-derived envelopes can serve as effective reward-shaping signals for improving contact transitions during locomotion.

\section{Conclusion}
We present ADAPT, a framework that augments RL-based humanoid control with an analytical whole-body disturbance observer. The observer provides a physics-grounded, sensorless force-related observation that can be directly fed into the policy as observation or used for reward shaping. Experiments on a Unitree G1 show that ADAPT improves robustness and velocity tracking under external forces, generalizes to OOD perturbations such as asymmetric hand loads, and enables emergent behaviors including light-footed locomotion through disturbance-derived reward shaping.

\section{Limitations}
The disturbance observer in ADAPT has a finite convergence time and thus responds with some delay to fast-varying disturbances such as impulsive pushes and foot-ground impacts. Currently, we absorb these fast transients into a single lumped disturbance, and ADAPT still exhibits robust performance in our experiments. However, this lumped treatment can make the learned policy somewhat conservative. Separating fast-varying components from the lumped signal, or designing observers for fast disturbances, could further improve responsiveness, which we leave for future work. ADAPT could also be combined with multimodal robot policy backbones~\citep{jia2026mars} to enable more expressive humanoid behaviors under disturbance-aware control.

\clearpage
\newpage
\bibliographystyle{assets/plainnat}
\bibliography{paper}

\clearpage
\newpage
\appendix

\section{Implementation details of disturbance observer}
\textbf{Parameters.}
The theoretical formulation of the disturbance observer is given in Sec.~\ref{sec:whole_body_observer}. Here, we report the parameters used for hardware deployment. The observer runs at $50~\mathrm{Hz}$, using the same update rate as the policy. The observer timestep is therefore $\Delta t_{\mathrm{obs}}=0.02~\mathrm{s}$. We use a uniform diagonal observer gain for all generalized velocity coordinates,
\begin{equation}
    K_o = k_o I_{6+n_j}, \qquad k_o = 3.0 ,
\end{equation}
where the first six dimensions correspond to the floating base and the remaining $n_j$ dimensions correspond to the actuated joints. All residual channels use the same observer gain in hardware experiments.

The raw residual estimation is smoothed before being used by the policy. We apply an online second-order Butterworth low-pass filter independently to each residual channel. The cutoff frequency is $f_c=1.0~\mathrm{Hz}$. The filtered residual is denoted by $\bar{\tau}_e$, consistent with Eq.~\eqref{eq:momo_tau_final}. It is then normalized and used as the disturbance observation for the policy.

\textbf{Approximation.}
\label{observer_approximation}
The momentum observer requires the bias term
$\beta(q,v)=C^\top(q,v)v-g(q)$. In the main implementation, we compute the observer with MuJoCo dynamics quantities for efficiency. MuJoCo provides the bias force
\begin{equation}
    q_{\mathrm{bias}} = C(q,v)v + g(q).
\end{equation}
We therefore approximate $\beta(q,v)$ with the negative MuJoCo bias force,
\begin{equation}
    \hat{\beta}_{\mathrm{mj}}(q,v)
    =
    -q_{\mathrm{bias}}
    =
    -C(q,v)v-g(q).
\end{equation}
Under the same convention used in Sec.~\ref{sec:whole_body_observer}, $\dot M(q)=C(q,v)+C^\top(q,v)$. The approximation error in $\beta$ is then
\begin{align}
    \epsilon_\beta(q,v)
    &=
    \beta(q,v)-\hat{\beta}_{\mathrm{mj}}(q,v) \\
    &=
    C^\top(q,v)v-g(q)+C(q,v)v+g(q) \\
    &=
    \dot M(q)v .
\end{align}
Thus, the MuJoCo approximation introduces only a velocity-dependent error in the observer bias term. This error is zero when \(v=0\).

We test the effect of this approximation with an independent Pinocchio-based observer. Pinocchio computes the exact term $\beta(q,v)=C^\top(q,v)v-g(q)$ from the full Coriolis matrix and the generalized gravity vector. The MuJoCo-based and Pinocchio-based observers use the same robot states, commanded torques, observer parameters, and filters. They differ only in the computation of $\beta(q,v)$.

During walking, we inject known joint-space torques into the left shoulder pitch, left elbow, right shoulder pitch, and waist pitch joints. These injected torques are used as ground truth. We then compare the residuals estimated by the two observers.

As shown in Fig.~\ref{fig:mujoco_pinocchio_comparison}, the two observers produce almost identical residuals on all tested joints. Both estimates match the sign, magnitude, and timing of the injected torques. The small differences near the rising and falling edges mainly come from the observer bandwidth and the output filter. This result shows that the velocity-dependent error in $\beta$ has little effect on the final disturbance estimate.

We use the MuJoCo-based observer in the main experiments to keep the computation consistent between policy training and deployment. We also implement the Pinocchio-based observer as an alternative backend for real-robot deployment.

\renewcommand*{\thefigure}{S1}
\begin{figure}
    \centering
    \includegraphics[width=0.8\linewidth]{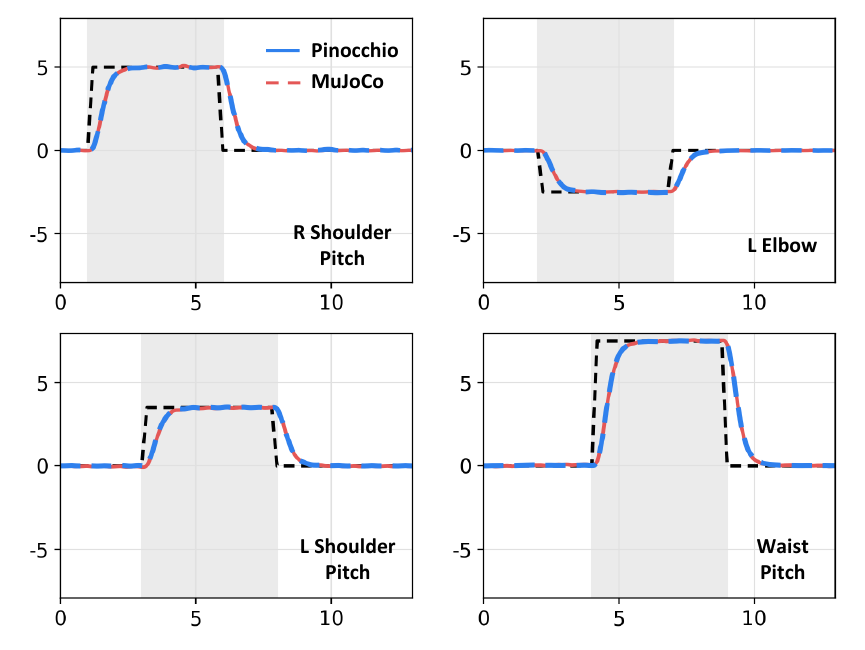}
    \caption{\textbf{Effect of approximation in observer.} Comparison between MuJoCo-based and Pinocchio-based observer residuals under known joint-space torque perturbations. The two observers produce nearly identical estimates and both match the injected torques, showing that the MuJoCo bias approximation has little practical effect.}
    \label{fig:mujoco_pinocchio_comparison}
\end{figure}

\section{Experiment setup}\label{experiment_setup}
\textbf{Training setup.}
All policies are trained in mjlab~\citep{zakka2026mjlablightweightframeworkgpuaccelerated} with the Unitree G1 model. Training is performed on an NVIDIA RTX 5090 GPU with $4096$ parallel environments. The simulator uses a physics timestep of $0.005~\mathrm{s}$ and a control decimation of $4$, resulting in a $200~\mathrm{Hz}$ simulation rate and a $50~\mathrm{Hz}$ policy update rate. The policy action is a joint-position offset added to the default pose and tracked by the low-level PD controller. We train ADAPT in two stages. Stage 1 trains the nominal walking policy for $15{,}000$ iterations. Stage 2 resumes from the Stage-1 checkpoint and trains the disturbance-aware policy for another $20{,}000$ iterations with external wrench perturbations applied on torso and end-effectors.

\textbf{Reward.}
The base reward is shared by ADAPT and the corresponding baselines. It consists of velocity tracking terms and regularization terms for posture, smoothness, contact quality, and safety. We denote the commanded velocity by $c=[v^{\mathrm{cmd}}_{x},v^{\mathrm{cmd}}_{y},\omega^{\mathrm{cmd}}_{z}]$, where $v^{\mathrm{cmd}}_{x}$ and $v^{\mathrm{cmd}}_{y}$ are the commanded base linear velocities and $\omega^{\mathrm{cmd}}_{z}$ is the commanded yaw rate. We use $v^b$ and $\omega^b$ for the base linear and angular velocities in the root frame, $q_{\mathrm{jnt}}$ for actuated joint positions, $\dot q_{\mathrm{jnt}}$ for actuated joint velocities, $a$ for policy actions, and $h_f$ for foot height. The reward terms are summarized in Table~\ref{tab:reward_terms}. The light-step reward is only used in the reward-shaping experiment.

The total reward is computed as a weighted sum,
\begin{equation}
    r = \sum_i w_i r_i ,
\end{equation}
where $r_i$ is the raw reward or penalty term and $w_i$ is its weight. For compact notation, let $\mathcal{F}$ denote the set of feet, $s_f\in{0,1}$ denote the contact state of foot $f$, $v_f$ denote the foot velocity, and $h_f^\star=0.1~\mathrm{m}$ denote the target foot height. We use $\chi_{\mathrm{cmd}}=\mathbb{I}(|[v_x^{\mathrm{cmd}},v_y^{\mathrm{cmd}}]|>0.05~\mathrm{or}~|\omega_z^{\mathrm{cmd}}|>0.05)$ to mask foot-related rewards when the robot is commanded to stand. $h_f^{\mathrm{peak}}$ denotes the maximum foot height reached during the preceding swing phase, and $\mathbb{I}^{\mathrm{fc}}_f$ indicates the first-contact event of foot $f$.

\renewcommand*{\thetable}{S1}
\begin{table}[t]
    \centering
    \caption{\textbf{Reward terms.}}
    \label{tab:reward_terms}
    \small
    \begin{adjustbox}{width=\linewidth}
    \begin{tabular}{l l c}
        \toprule
        Term & Raw term $r_i$ & $w_i$ \\
        \midrule
        Linear velocity tracking
        & $\exp\!\left(-\frac{\|v^b_{xy}-[v_x^{\mathrm{cmd}},v_y^{\mathrm{cmd}}]\|^2}{0.25}\right)$
        & $2.0$ \\

        Yaw-rate tracking
        & $\exp\!\left(-\frac{(\omega^b_z-\omega_z^{\mathrm{cmd}})^2}{0.25}\right)$
        & $3.0$ \\

        Vertical velocity
        & $(v^b_z)^2$
        & $-2.0$ \\

        Upright torso
        & $\exp\!\left(-\frac{\|g^b_{xy}\|^2}{0.2}\right)$
        & $1.0$ \\

        Joint posture
        & $\exp\!\left(-\left\|\frac{q_{\mathrm{jnt}}-q_{\mathrm{jnt}}^{\mathrm{def}}}{\sigma_q}\right\|^2\right)$
        & $1.0$ \\
        
        Joint position limits
        & $\sum_{i=1}^{n_j}\left[\max(q_{\mathrm{jnt},i}-q^{\max}_{\mathrm{jnt},i},0)^2+\max(q^{\min}_{\mathrm{jnt},i}-q_{\mathrm{jnt},i},0)^2\right]$
        & $-1.0$ \\

        Action rate
        & $\|a-a_{\mathrm{prev}}\|^2$
        & $-0.1$ \\

        Foot clearance
        & $\chi_{\mathrm{cmd}}\sum_{f\in\mathcal{F}}\|v_{f,xy}\|_2\,|h_f-h_f^\star|$
        & $-2.0$ \\
    
        Swing-foot height
        & $\chi_{\mathrm{cmd}}\sum_{f\in\mathcal{F}}\mathbb{I}^{\mathrm{fc}}_f
        \left(\frac{h_f^{\mathrm{peak}}}{h_f^\star}-1\right)^2$
        & $-0.25$ \\

        Foot slip
        & $\chi_{\mathrm{cmd}}\sum_{f\in\mathcal{F}}s_f\|v_{f,xy}\|^2$
        & $-0.1$ \\

        Soft landing
        & $\chi_{\mathrm{cmd}}\sum_{f\in\mathcal{F}}\mathbb{I}(s_f=1)\|v_{f,z}\|^2$
        & $-1.0\times10^{-5}$ \\

        Torso angular velocity
        & $\|\omega^b_{xy}\|^2$
        & $-0.05$ \\

        Angular momentum
        & $\|L_{\mathrm{root}}\|^2$
        & $-0.02$ \\

        Self-collision
        & $\mathbb{I}_{\mathrm{self}}$
        & $-1.0$ \\

        Light-step reward
        & Eq.~\eqref{eq:light_step_reward}
        & experiment-specific \\
        \bottomrule
    \end{tabular}
    \end{adjustbox}
\end{table}

\textbf{Domain randomization.}
We apply domain randomization to improve sim-to-real transfer. The randomized quantities are listed in Table~\ref{tab:domain_randomization}. Here, $\mu$ denotes the ground friction coefficient, $b_q$ denotes encoder bias, $\Delta p_{\mathrm{com}}$ denotes the torso center-of-mass offset, $\alpha_{\mathrm{pi}}$ denotes the pseudo-inertia scaling factor, $\tau_{\mathrm{fric}}^{\mathrm{rand}}$ denotes the randomized joint friction term, and $s_d$ denotes the joint damping multiplier. The pseudo-inertia randomization scales mass and inertia consistently for the leg, torso, and arm links. For joint friction and damping, the ranges in the table denote the final values reached by the curriculum during Stage 2.

\renewcommand*{\thetable}{S2}
\begin{table}[t]
    \centering
    \caption{\textbf{Domain randomization.}}
    \label{tab:domain_randomization}
    \footnotesize
    \setlength{\tabcolsep}{4pt}
    \begin{adjustbox}{max width=\linewidth}
    \begin{tabular}{@{}l l l@{}}
        \toprule
        Term & Description & Range / setting \\
        \midrule
        $\mu$ & Ground friction coefficient & $[0.3,1.2]$ \\
        $b_q$ & Encoder bias & $[-0.015,0.015]~\mathrm{rad}$ \\
        $\Delta p_{\mathrm{com},x}$ & Torso COM offset along $x$ & $[-0.025,0.025]~\mathrm{m}$ \\
        $\Delta p_{\mathrm{com},y}$ & Torso COM offset along $y$ & $[-0.025,0.025]~\mathrm{m}$ \\
        $\Delta p_{\mathrm{com},z}$ & Torso COM offset along $z$ & $[-0.03,0.03]~\mathrm{m}$ \\
        $\alpha_{\mathrm{pi}}$ & Pseudo-inertia scaling factor & $[-0.05,0.05]$ \\
        $\tau^{\mathrm{rand}}_{\mathrm{fric},\mathrm{leg}}$ & Leg joint friction & $[0.05,1.5]$ \\
        $\tau^{\mathrm{rand}}_{\mathrm{fric},\mathrm{waist}}$ & Waist joint friction & $[0.05,1.0]$ \\
        $\tau^{\mathrm{rand}}_{\mathrm{fric},\mathrm{arm}}$ & Arm joint friction & $[0.05,0.8]$ \\
        $s_{d,\mathrm{leg}}$ & Leg joint damping scale & $[0.5,1.5]$ \\
        $s_{d,\mathrm{waist}}$ & Waist joint damping scale & $[0.7,1.3]$ \\
        $s_{d,\mathrm{arm}}$ & Arm joint damping scale & $[0.7,1.3]$ \\
        $\epsilon_{v^b}$ & Base linear velocity noise & $[-0.5,0.5]~\mathrm{m/s}$ \\
        $\epsilon_{\omega^b}$ & Base angular velocity noise & $[-0.2,0.2]~\mathrm{rad/s}$ \\
        $\epsilon_{g^b}$ & Projected gravity noise & $[-0.05,0.05]$ \\
        $\epsilon_{q_{\mathrm{jnt}}}$ & Joint position noise & $[-10^{-4},10^{-4}]~\mathrm{rad}$ \\
        $\epsilon_{\dot q_{\mathrm{jnt}}}$ & Joint velocity noise & $[-0.03,0.03]~\mathrm{rad/s}$ \\
        \bottomrule
    \end{tabular}
    \end{adjustbox}
\end{table}

\textbf{Deployment setup.}
All real-world experiments are conducted on the Unitree G1 humanoid. The high-level controller runs on an external laptop with an NVIDIA RTX 5070 Ti Laptop GPU. The laptop is connected to the robot through wired Ethernet. It receives joint positions, joint velocities, IMU measurements, and motor command information from the robot. It sends joint-position targets back to the robot. The onboard low-level controller tracks these targets with PD control.

FAST-LIO~\citep{xu2022fast} runs on the onboard computer of the G1. It publishes odometry at $100~\mathrm{Hz}$. The odometry provides velocity estimation and global localization. The laptop uses the latest FAST-LIO message to compute the root linear velocity, following $v^{R}_{\mathrm{root}} = v^{R}_{L} - J_{L}(q_{\mathrm{jnt}})\dot{q}_{\mathrm{jnt}}$. This velocity is used by both the policy observation and the disturbance observer.

The policy and the disturbance observer run on the laptop at $50~\mathrm{Hz}$, matching the policy update rate used during training. At each control step, the laptop first builds the current robot state from proprioception and FAST-LIO odometry. It then writes this state and the latest motor command into a local MuJoCo model. MuJoCo is not used to simulate future robot motion during deployment. Instead, it serves as a synchronized nominal dynamics model. After state synchronization, we call MuJoCo forward dynamics to compute the nominal dynamics terms required by the momentum observer. The observer updates the filtered residual using these nominal dynamics terms and the measured robot state. The filtered residual is normalized and concatenated with the policy observation. The policy then outputs the next joint-position target for onboard PD tracking.

\section{Task descriptions.}

\textbf{Torso pulling test.} We evaluate torso-level disturbance rejection in simulation by applying constant horizontal forces to the robot torso while it tracks a fixed forward velocity command. Each rollout starts from the reset state, followed by a $1.0\,\mathrm{s}$ settling phase with zero command and zero external force. The robot is then commanded with $\mathbf{v}^{\mathrm{cmd}}=[v_x^{\mathrm{cmd}},0,0]$ for $10.0\,\mathrm{s}$, while a constant world-frame force is applied to \texttt{torso\_link}. The applied wrench has zero moment and zero vertical-force component. For the sagittal-force evaluation, we set $v_x^{\mathrm{cmd}}\in\{0.5,0.8,1.1\}\,\mathrm{m/s}$ and apply $F_x\in\{-60,-40,-20,0,20,40,60\}\,\mathrm{N}$ with $F_y=0$. For the directional evaluation, we fix $v_x^{\mathrm{cmd}}=1.1\,\mathrm{m/s}$ and apply forces with magnitudes of $40\,\mathrm{N}$ and $60\,\mathrm{N}$ along eight horizontal directions. Each simulated condition is evaluated with ten random seeds, and rollouts are recorded at the policy control rate.

\textbf{Asymmetric hand loading test.} We evaluate asymmetric loading by applying a unilateral downward load to the right hand. In simulation, the load is modeled as a constant force applied to \texttt{right\_wrist\_yaw\_link}. The commanded velocity is $\mathbf{v}^{\mathrm{cmd}}=[v_x^{\mathrm{cmd}},0,0]$, where $v_x^{\mathrm{cmd}}\in\{0.5,0.8,1.1\}\,\mathrm{m/s}$, and the downward force magnitude is varied over $\{10,20,40\}\,\mathrm{N}$. Each condition is evaluated with ten random seeds for both the baseline policy and ADAPT. For the real-robot test, the load is physically attached to the right hand. We use payloads of $1\,\mathrm{kg}$, $2\,\mathrm{kg}$, and $4\,\mathrm{kg}$ with forward commands of $0.8\,\mathrm{m/s}$ and $1.1\,\mathrm{m/s}$. The reported real-world comparison uses three repeated trials per policy for each command-load condition.

\textbf{Light-step reward shaping.} We evaluate light-step reward shaping during forward walking, without external payloads or applied disturbances. The robot tracks a fixed command $\mathbf{v}^{\mathrm{cmd}}=[0.8,0,0]$. In simulation, we compare the baseline policy and the policy trained with the light-step reward over $20$ random seeds per policy. For each trajectory, the observer is already running before the walking command starts. We then collect command-following data to compute the reported curves. For the real-robot evaluation, we use the same forward command and compare the baseline policy with the reward-shaped policy. Since the real robot does not provide ground-truth foot contact forces, we report the observer-based leg disturbance envelope instead of measured contact force. The envelope is computed from the normalized residuals of the $12$ leg joints. We first compute the envelope for each leg and then take the maximum over the left and right leg envelopes. The reported real-world comparison uses seven completed trials per policy.

\section{Evaluation criterion}
\label{Evaluation_criterion}
We report three metrics for fixed-command locomotion. The forward \textbf{velocity tracking error} is computed as the root-mean-square error between the commanded forward velocity $v_x^{\mathrm{cmd}}$ and the forward velocity $v^b_{x,k}$ of root in over the evaluation window $\mathcal{T}$
\begin{equation}
    e_{v_x} =
    \sqrt{
    \frac{1}{|\mathcal{T}|}
    \sum_{k \in \mathcal{T}}
    \left(v^b_{x,k} - v_x^{\mathrm{cmd}}\right)^2
    } .
\end{equation}
In forward-velocity tracking tasks, the robot is expected not only to maintain the commanded speed but also to preserve its walking direction. We therefore evaluate directional stability using two metrics: \textbf{lateral drift} and \textbf{yaw drift}
\begin{equation}
    d_y = y_{k_{\mathrm{end}}} - y_{k_{\mathrm{start}}},
\end{equation}
where the position is measured in the world frame, and $k_{\mathrm{start}}$ and $k_{\mathrm{end}}$ denote the first and last timesteps of the evaluation window. The yaw drift is computed as the wrapped heading change,
\begin{equation}
    d_{\psi} =
    \mathrm{wrap}_{[-\pi,\pi]}
    \left(\psi_{k_{\mathrm{end}}} - \psi_{k_{\mathrm{start}}}\right).
\end{equation}
For policy comparison, we report the magnitudes $|d_y|$ and $|d_{\psi}|$, where lower values indicate better rejection of lateral and rotational deviations under external disturbances or asymmetric payloads.

\section{Unified-scale ablation}
\label{effort_scale_ablation}
We evaluate unified-scaled normalization for two stages in velocity tracking under external force perturbations as the baseline. We compare ADAPT with a standard-normalization variant, where the estimated disturbances are normalized by standard statistics instead of a unified normalization. As shown in Fig.~\ref{fig:ablation_effort_scale}, ADAPT tracks the velocity more robustly under perturbations with lower drift. This result shows that joint effort provides a useful scale for representing disturbance estimates.

\renewcommand*{\thefigure}{S2}
\begin{figure*}
    \centering
    \includegraphics[width=1\linewidth]{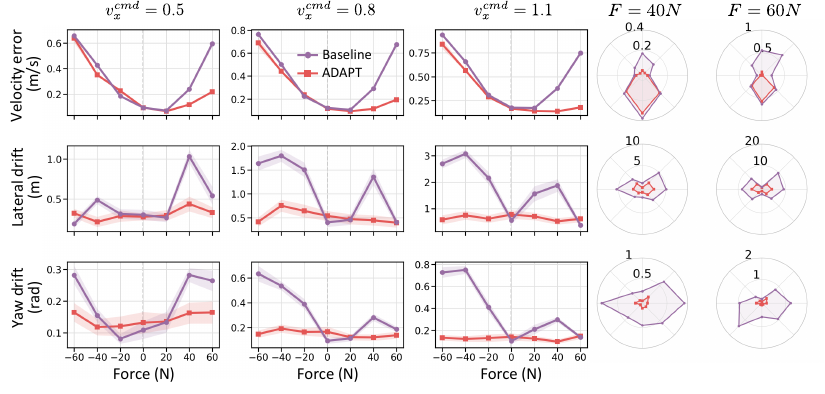}
    \caption{\textbf{Unified-scale normalization ablation.} The baseline here denotes a variant of ADAPT using standard-normalization instead of our unified normalization.
    \textit{Left:} Velocity tracking error, lateral drift, and yaw drift under sagittal torso forces across three command velocities. \textit{Right:} Radar plots summarize results across disturbance directions.}
    \label{fig:ablation_effort_scale}
\end{figure*}

\section{Supplementary experiment results}
\textbf{Asymmetric hand loading test.} Fig.~\ref{fig:appendix_full_handload} provides additional results for the asymmetric hand loading test. In simulation, we apply one-sided downward forces on the robot hand with magnitudes of $10~\mathrm{N}$, $20~\mathrm{N}$, and $40~\mathrm{N}$. In the real-world experiment, we attach asymmetric payloads to one hand, including $1~\mathrm{kg}$, $2~\mathrm{kg}$ and $4~\mathrm{kg}$ loads. For both simulation and real-world tests, we evaluate fixed-forward-velocity walking under two commands, $v_x^{\mathrm{cmd}}=0.8~\mathrm{m/s}$ and $v_x^{\mathrm{cmd}}=1.1~\mathrm{m/s}$, the supplementary comparison reports the matching simulation results under the same two forward velocity commands.

The plots show the lateral drift with respect to the forward walking distance. Across both simulation and real-world experiments, ADAPT produces smaller lateral drift than the baseline under asymmetric loading. This indicates that ADAPT can better maintain the commanded walking direction under unbalanced disturbances.

\renewcommand*{\thefigure}{S3}
\begin{figure*}
    \centering
    \includegraphics[width=1\linewidth]{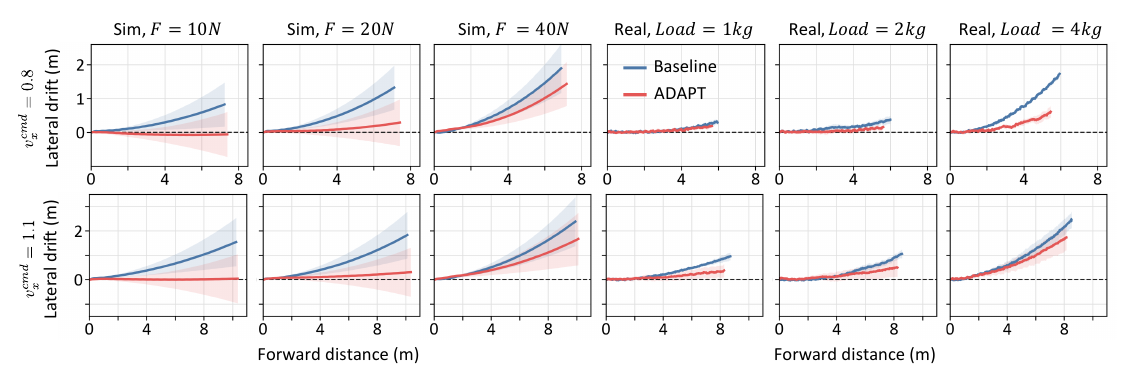}
    \caption{\textbf{Supplementary results for the asymmetric hand loading test.}
    We report lateral drift trajectories under one-sided hand loads in simulation and on the real robot. ADAPT maintains a straighter walking direction than the baseline under asymmetric disturbances.}
    \label{fig:appendix_full_handload}
\end{figure*}

\end{document}